# Capacity limitations of visual search in deep convolutional neural networks

Endel Põder

Institute of Psychology, University of Tartu, Estonia

## Abstract

Deep convolutional neural networks follow roughly the architecture of biological visual systems and have shown a performance comparable to human observers in object recognition tasks. In this study, I tested three pretrained deep neural networks in visual search for simple visual features, and for feature configurations. The results reveal a qualitative difference from human performance. It appears that there is no clear difference between searches for simple features that pop out in experiments with humans, and for feature configurations that exhibit strict capacity limitations in human vision. Both types of stimuli reveal comparable capacity limitations in the neural networks tested here.

## Introduction

It is well known that human observers have certain limitations on simultaneous processing of multiple visual stimuli (Estes & Taylor, 1964; Bergen & Julesz, 1983). Visual search experiments have revealed several simple features (luminance, color, size, orientation) that can be detected in parallel across the visual field, independent of the number of objects (e.g. Wolfe, 1998). Detection of combinations of simple features is more difficult and may need serial processing (Treisman & Gelade, 1980; Wolfe et al, 1989). Signal detection theory that assumes noisy representation of feature values has slightly changed the picture (Kinchla, 1974; Palmer et al, 2000), but different behavior of simple and complex features is still important. Search for a simple feature among homogeneous distractors fits well to SDT model that assumes independent encoding of visual objects, and ideal integration of noisy signals (Shaw, 1984; Palmer et al, 1993; Palmer, 1994). Search for configurations of simple features has strictly limited capacity and exhibit set size effects consistent with dividing fixed processing resources, or serial scanning (Shaw, 1984; Põder, 1999; Palmer et al, 2011).

According to a widely accepted view, spatial attention plays an important role in perception of complex objects (Treisman & Gelade, 1980: Cheal et al, 1991; Wolfe & Bennett, 1996). It is believed that spatial attention gates visual signals at relatively low levels and in retinotopic coordinates and thus simplifies processing at higher levels (Broadbent, 1958; Neisser, 1967).

However, there are different opinions too (Deutsch & Deutsch, 1963; Allport, Tipper & Chmiel, 1985). In recent studies, Rosenholtz (Rosenholtz et al 2012; Rosenholtz, 2017) has argued that spatial gating is not necessary in visual processing and apparent capacity limitations may reflect complexity of decision boundary in some high-level multidimensional space where a representation of the whole visual field is encoded. This view resembles information processing in current artificial neural networks. Therefore, experimenting with these networks may help to test theories of biological vision.

Several years ago, artificial neural networks reached the level of human performance in demanding visual object recognition tasks (e.g. Krizhevsky et al, 2012; Ciresan et al, 2012; Simonyan & Zisserman, 2014; Szegedy et al, 2015). These networks are hierarchical feature combiners following roughly the architecture of biological visual systems and trained on millions of labeled natural images.

Many studies have reported on functional similarities between deep neural networks and visual systems of humans or monkeys (e.g. Khaligh-Razavi & Kriegeskorte, 2014; Yamins et al, 2014; Kubilius et al, 2016). However, some interesting differences have been reported as well (Nguyen et al., 2015; Geirhos et al., 2018; Lonnqvist et al, 2020).

Up to now, there have been no publications on classic visual search experiments with deep neural networks. Usually, these networks do not contain any mechanisms of spatial attention. Therefore, it would be interesting to see whether they are able to reproduce the capacity limitations found in experiments with humans. In this study, I run simple search experiments using three well-known convolutional neural networks in place of a human observer.

# Methods

In the present experiments, the pretrained neural networks AlexNet, GoogLeNet, and ResNet18 provided with Matlab Deep Learning Toolbox were used. The last three layers that were adapted for the classification of 1000 natural image categories were removed and replaced with equivalent layers for the classification into two categories: "target present", and "target absent". Only one of the new layers in each network contained trainable weights and biases. These parameters were adjusted during the training with my visual search stimuli. The changes in the previous layers were prohibited by freezeWeights function.

Simple search stimuli were generated in Matlab. The size of stimuli was 227x227 or 224x224 pixels x 3 color planes. Each image contained n (n = 1, 2, 4, or 8) simple items (squares, lines, rectangles, rotated Ts). The items were depicted on a dark background. To minimize possible crowding effects, the minimal center-to-center distance between the items was set to be at least 48 pixels. Also, the items were not placed within 28-pixel edges of the image. Otherwise, the items were located randomly. The images of "target present" category contained one target item and n-1 distractor items, the images of "target absent" category contained only n distractors.

In this study, five search experiments with different visual features were run (examples of stimuli are given in Figure 1). There were four "simple" tasks, with targets of either different luminance, color, length, or orientation, and one "complex" task (rotated Ts), where target differs from distractors by spatial configuration of two bars. In addition to set size, difficulty levels were varied by either target-distractor difference, or size of stimuli.

Two different training procedures, with mixed and separate set sizes, were run.

For a mixed training, 8000 images were used, 4000 of "target present", and 4000 of "target absent" category. Each set size (1, 2, 4, and 8) had equal number of samples in both categories. 6400 images were assigned to training and 1600 to validation set. After training in a given search task, the network was tested with the same task using an independent sample (400) of images for each set size.

In separate training, total numbers of images were identical. In a single training run, there were 2000 images, 1000 of "target present", and 1000 of "target absent" category, 1600 images were assigned to training, and 400 to validation set. After training, the final proportion correct for the validation set was determined.

Training, with stochastic gradient descent, and constant training rate, was run on a single GPU. Detailed training parameters were somewhat different across the networks (minibatch size 50 or 80, training epochs 20 or 30, learning rate from 0.0003 to 0.003).

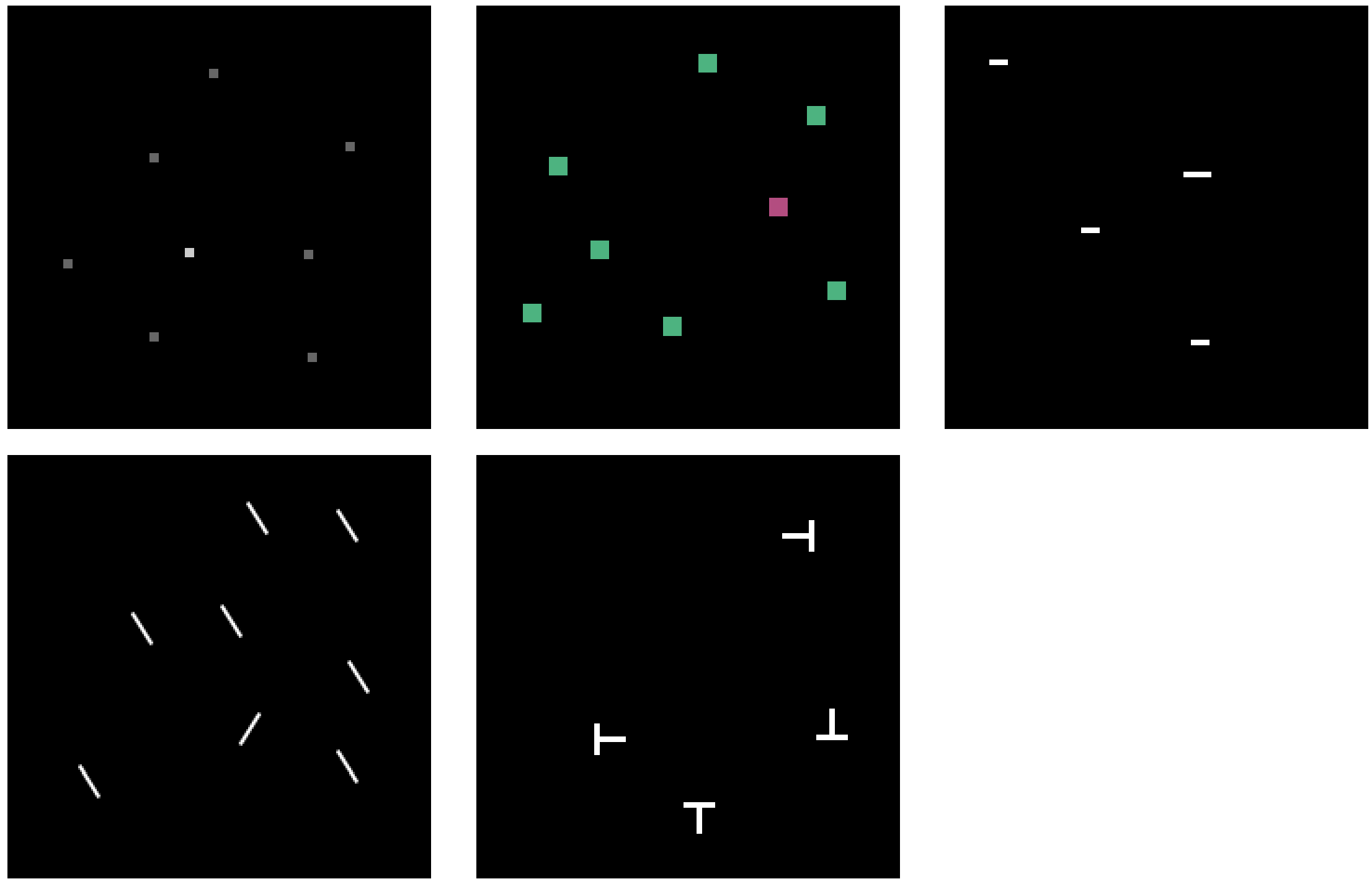

Figure 1. Examples of visual search stimuli used in this study. All examples depict stimuli with target present. Set size (number of objects in a display) was varied from 1 to 8.

Two simple models were used to measure the effects of set size on proportion correct. The first one supposes that d prime vs. set size slopes are constant in log-log graphs and measures this slope. The second is an SDT based search model with (possibly) limited capacity encoding and ideal decision rule (Palmer et al, 2000; Mazyar et al, 2012; Põder, 2017) that has been frequently applied to human observers. This model measures an effect of set size on encoding precision (noise variance). This measure is 0 for unlimited capacity (independent processing of items), and 1 for a fixed capacity (noise variance inversely proportional to set size). The two models have equal numbers of free parameters and are easily comparable.

## Results

Examples of data from individual experiments and model fits are depicted in Figure 2. The fits were far from perfect, but both models capture the main regularities of data reasonably well. Goodness-of-fit statistics for both models are given is Tables 1, and 2, for mixed and separate training, respectively. These results reveal consistently better fit for the results from mixed training experiments. The reason for that is not clear. A somewhat better fit of SDT based search model compared to the simple log-log slope model was observed as well.

The estimated measures of set size effects are given in Figures 3 and 4. Most interestingly, there is no systematic difference between classic simple features and complex feature configuration (rotated T) search. In both conditions, neural networks exhibit from moderate to strong capacity limitations. Still, there are clear difference between the networks – AlexNet has stronger capacity limitations as compared to GoogLeNet and ResNet18.

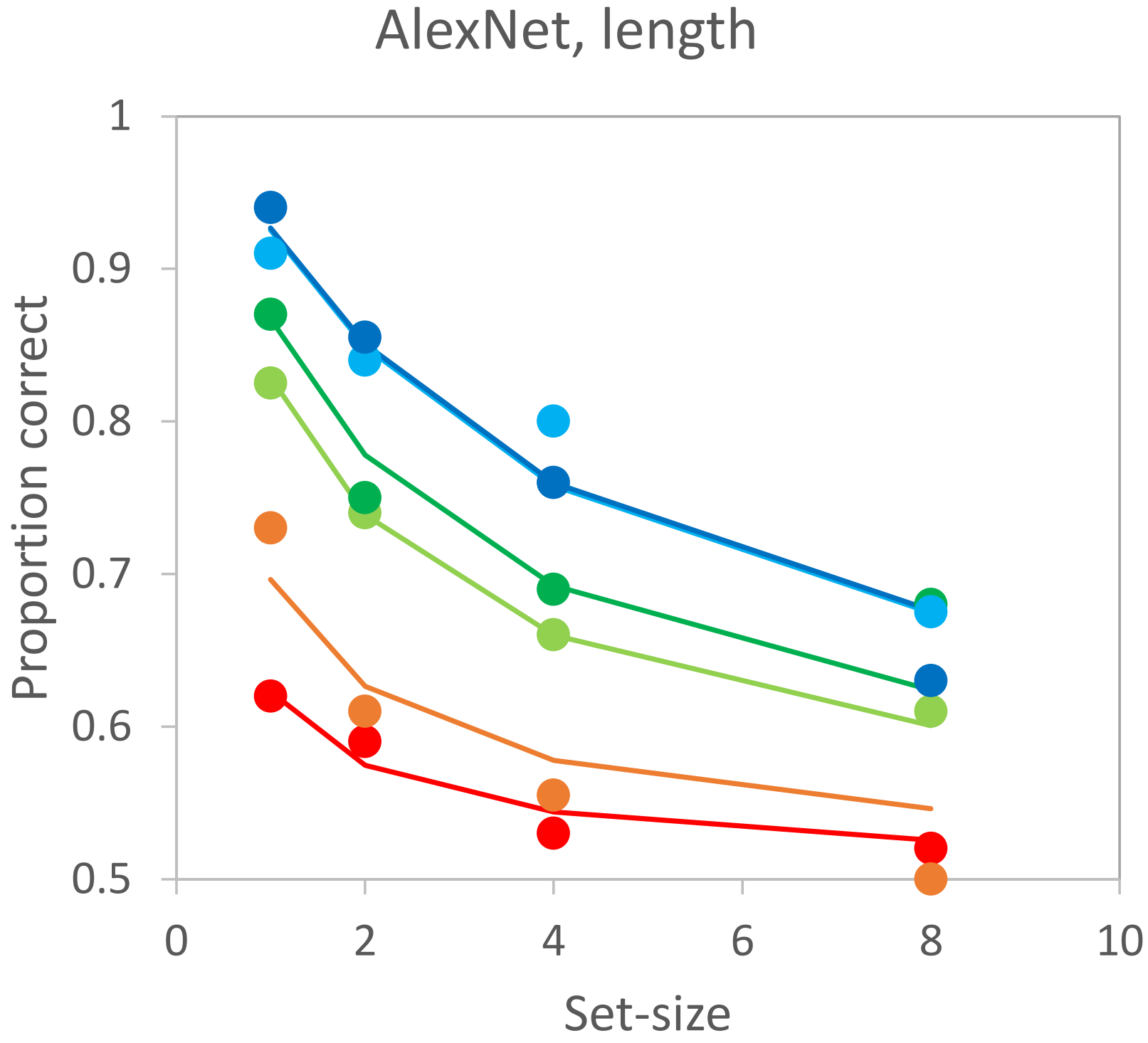

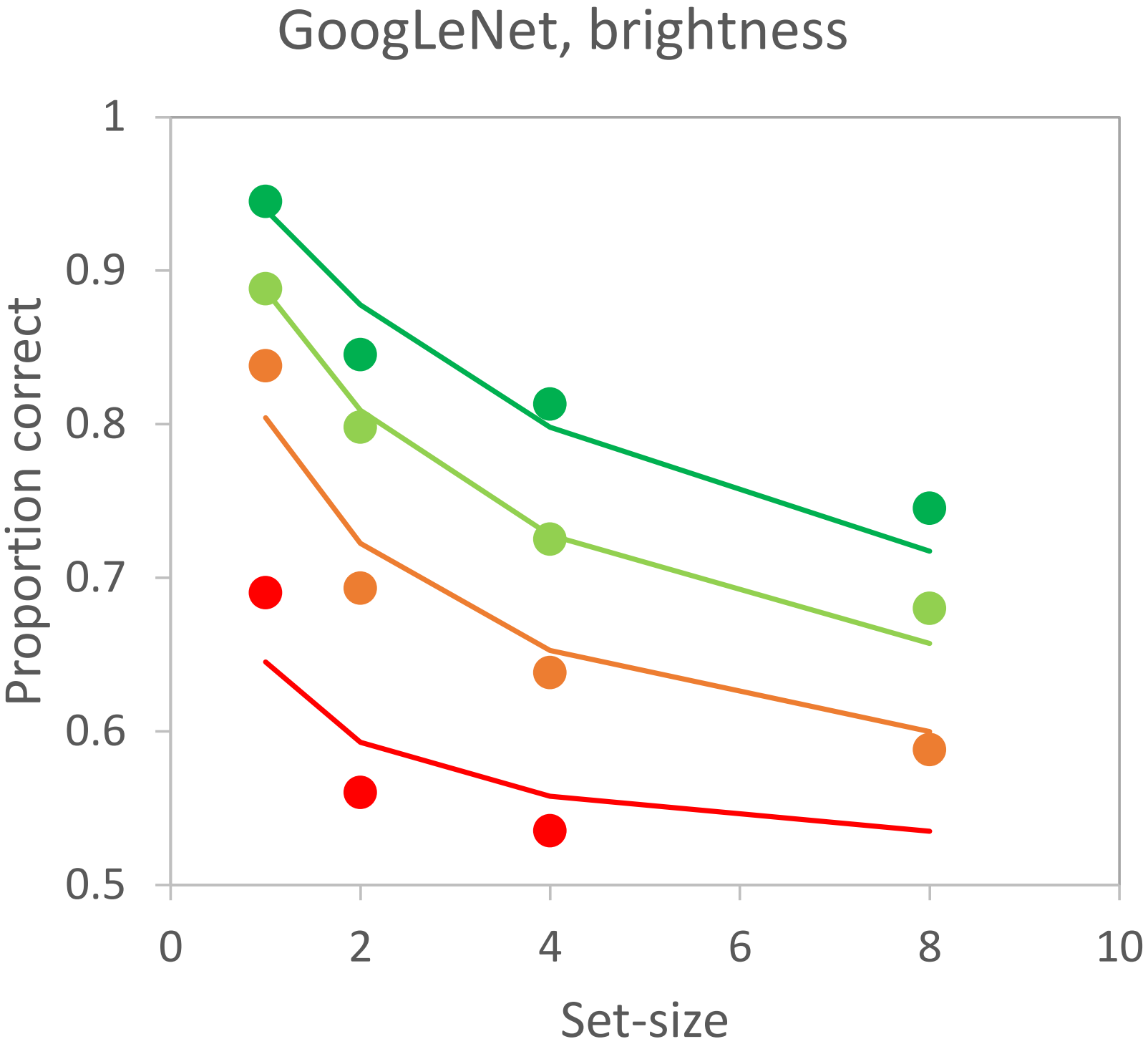

Figure 2. Examples of graphs with proportions correct as dependent on set-size and target-distractor discriminability, for different search experiments. Symbols depict experimental data and lines are model fits.

| | Constant log-log slope | | | SDT model | | |
|---|---|---|---|---|---|---|
| | AlexNet | GoogLeNet | ResNet18 | AlexNet | GoogLeNet | ResNet18 |
| Brightness | 22.0 | 28.8* | 6.5 | 22.0 | 22.1 | 6.9 |
| Length | 31.1 | 22.1 | 11 | 25.7 | 18.1 | 14.3 |
| Color | 95.2* | 10.5 | 44.4* | 72.5* | 15.1 | 11.1 |
| Orientation | 46.4* | 21.8 | 23.6 | 42.5* | 21 | 13.8 |
| Rotated T1 | 29.3* | 13.2 | 14.9 | 30.5* | 12.6 | 10.8 |
| Rotated T2 | 8.3 | 20.9 | 23.9 | 7.9 | 29.3* | 15.5 |
| Rotated T3 | 11.8 | 24.2 | 12 | 10.6 | 26.1 | 11.2 |
| Rotated T4 | 22.1 | 26.8 | 17.2 | 21.3 | 25.8 | 11 |

Table 1. Model fits (likelihood ratio statistic G) of the results from search experiments with mixed training. Significant differences (with $p<0.01$) between data and models are indicated.

| | Constant log-log slope | | | SDT model | | |
|---|---|---|---|---|---|---|
| | AlexNet | GoogLeNet | ResNet18 | AlexNet | GoogLeNet | ResNet18 |
| Brightness | 26.0* | 23.6 | 19.1 | 19.9 | 18.1 | 7.5 |
| Length | 44.8* | 40.8* | 38.9* | 38.3* | 39.2* | 28.9 |
| Color | 169.6* | 13.5 | 14.9 | 137.7* | 18.1 | 6.5 |
| Orientation | 53.3* | 30.5* | 37.2* | 42.5* | 31.1* | 30.8* |
| Rotated T1 | 53.3* | 51.0* | 31.7* | 41.6* | 41.1* | 21.1 |
| Rotated T2 | 34.5* | 38.1* | 47.5* | 29.4* | 41.4* | 39.5* |
| Rotated T3 | 21.4 | 31.2* | 33.2* | 22.8 | 33.9* | 25.4 |
| Rotated T4 | 18.0 | 15.6 | 33.2* | 16.2 | 12.8 | 25.5 |

Table 2. Model fits (likelihood ratio statistic G) of the results from search experiments with separate training. Significant differences (with $p<0.01$) between data and models are indicated.

# Discussion

In the present study, simple visual search experiments were run on pretrained deep convolutional neural networks. While many previous studies with human observers have found big differences between searches for simple features and for feature configurations, virtually no difference was found with artificial neural networks. Both types of stimuli revealed moderate to strong capacity limitations in the studied neural networks.

Quantitatively, d prime decreased with increasing set size with an exponent of -0.5 to -1. According to SDT based search model, encoding noise variance increased with set size according to a similar slope. Human observers, when searching for simple visual features, usually exhibit very small set size effects, and zero capacity limitations according to SDT model.

There is no question that deep convolutional networks can learn to accomplish these simple tasks much better, when allowed to adapt weights in the lower layers. However, the purpose of this study was to examine how well the learned image transformations necessary for object recognition support visual search. Apparently not very well.

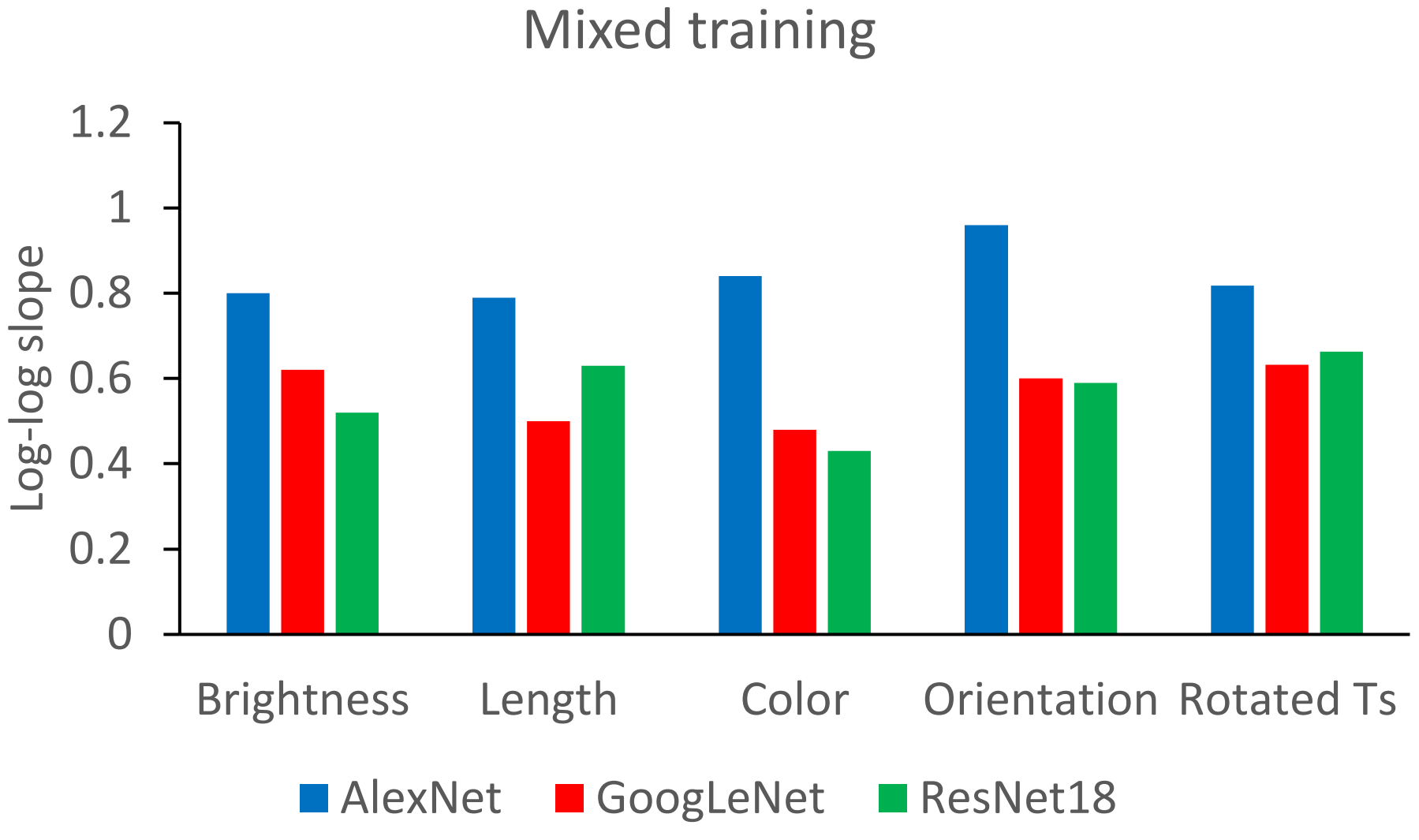

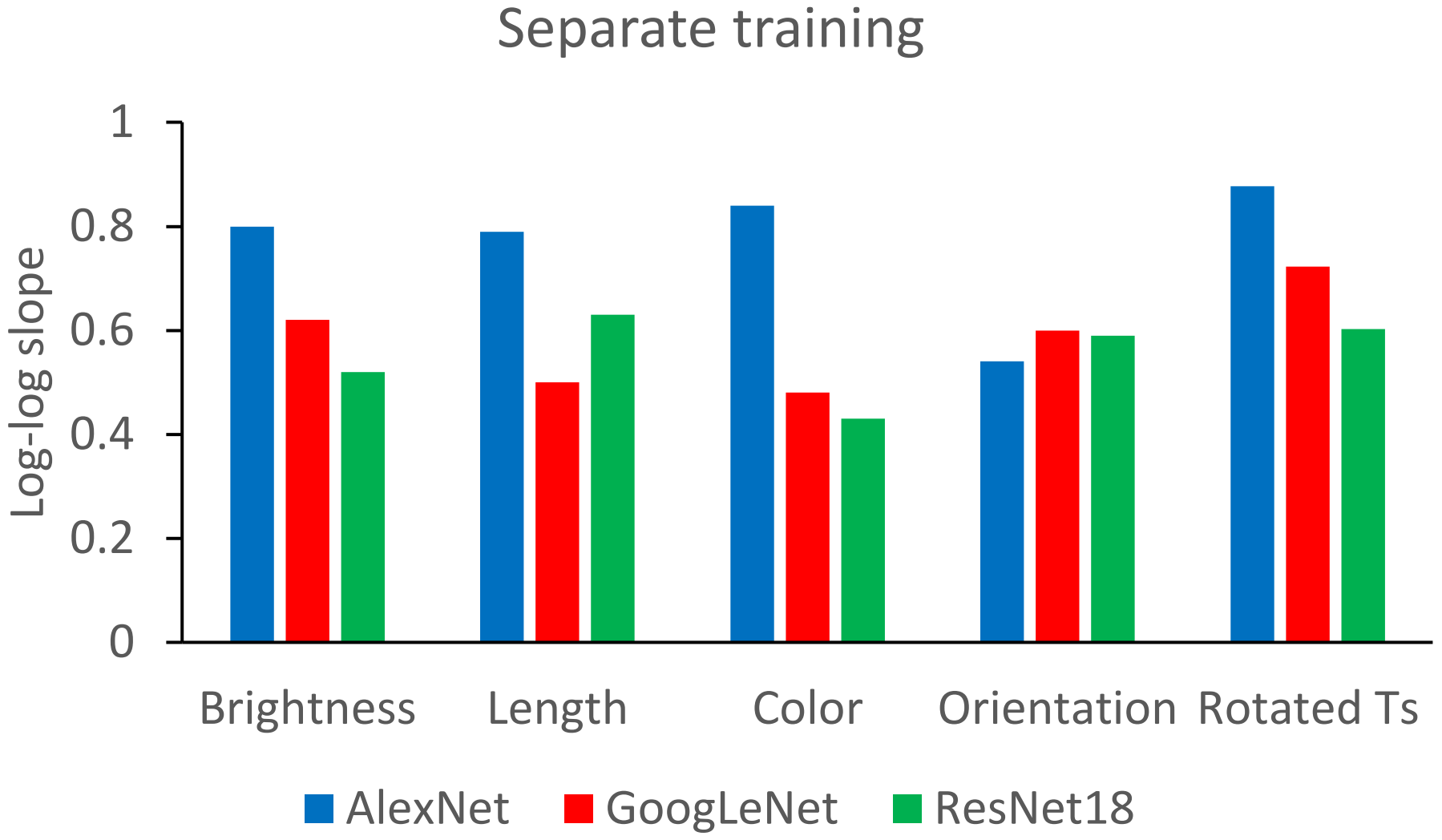

Figure 3. Set size effects as log-log slopes of d prime vs. set size for different search experiments, different neural networks, and for two training procedures.

There are several potential explanations. It is possible that artificial neural networks can acquire more human-like capacity limits when trained on more heterogeneous stimuli and with different visual tasks. However, some details of network architecture may be important too.

To understand this issue better, I tried to train a couple of toy networks. A simple 3-layer network, trained from scratch, learns easily to detect a target – a spike with a higher magnitude among distractors with a lower magnitude. However, the performance depends heavily on the number of spikes, very similarly to the set size effects observed in transfer learning with deep networks. Some prewired details, for example, a global max pooling layer, can help network to find a much more efficient algorithm, with zero set size effect. Therefore, it is possible that efficient performance of simple search tasks requires some built-in details not present in standard networks optimized for object classification.

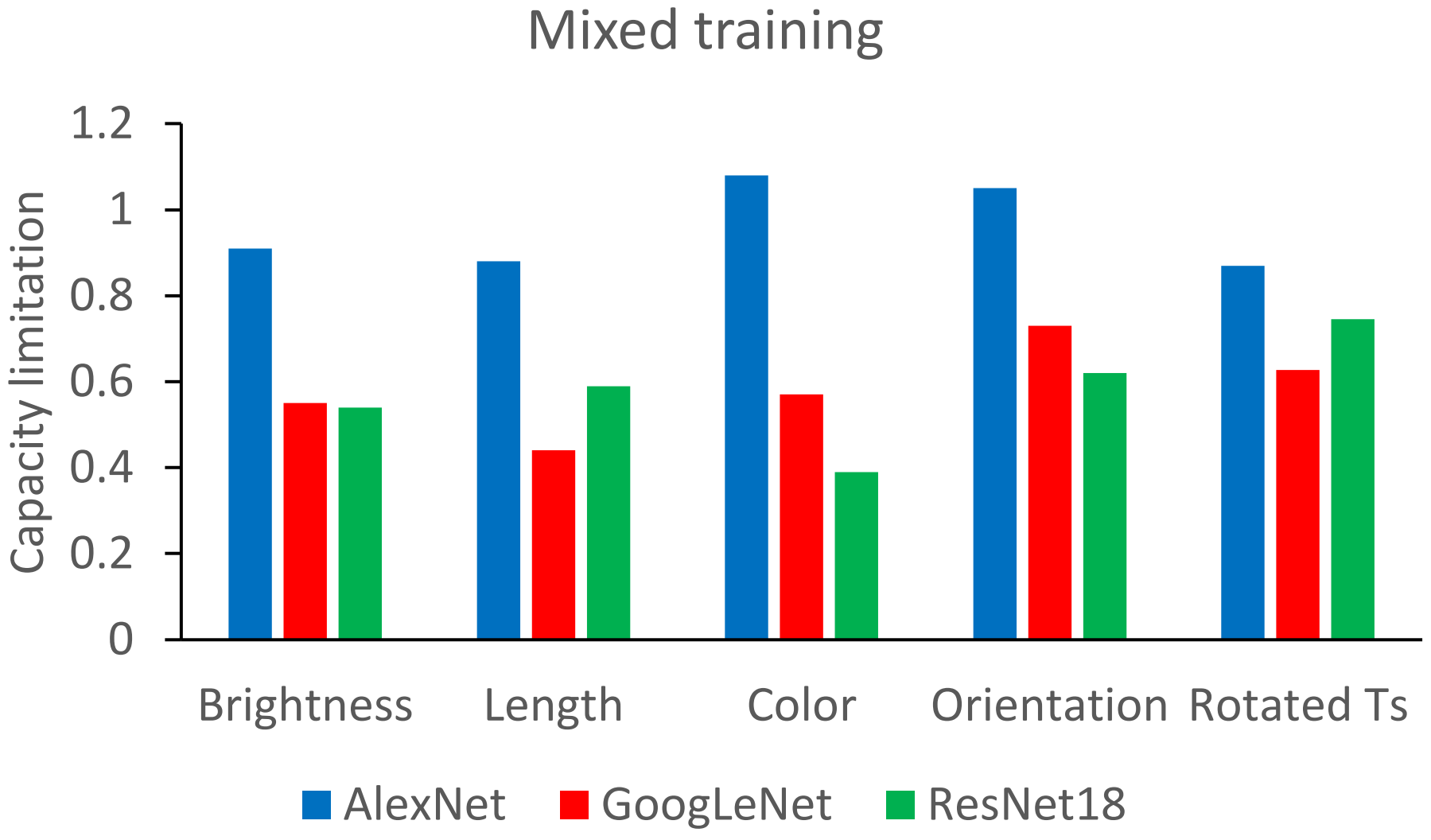

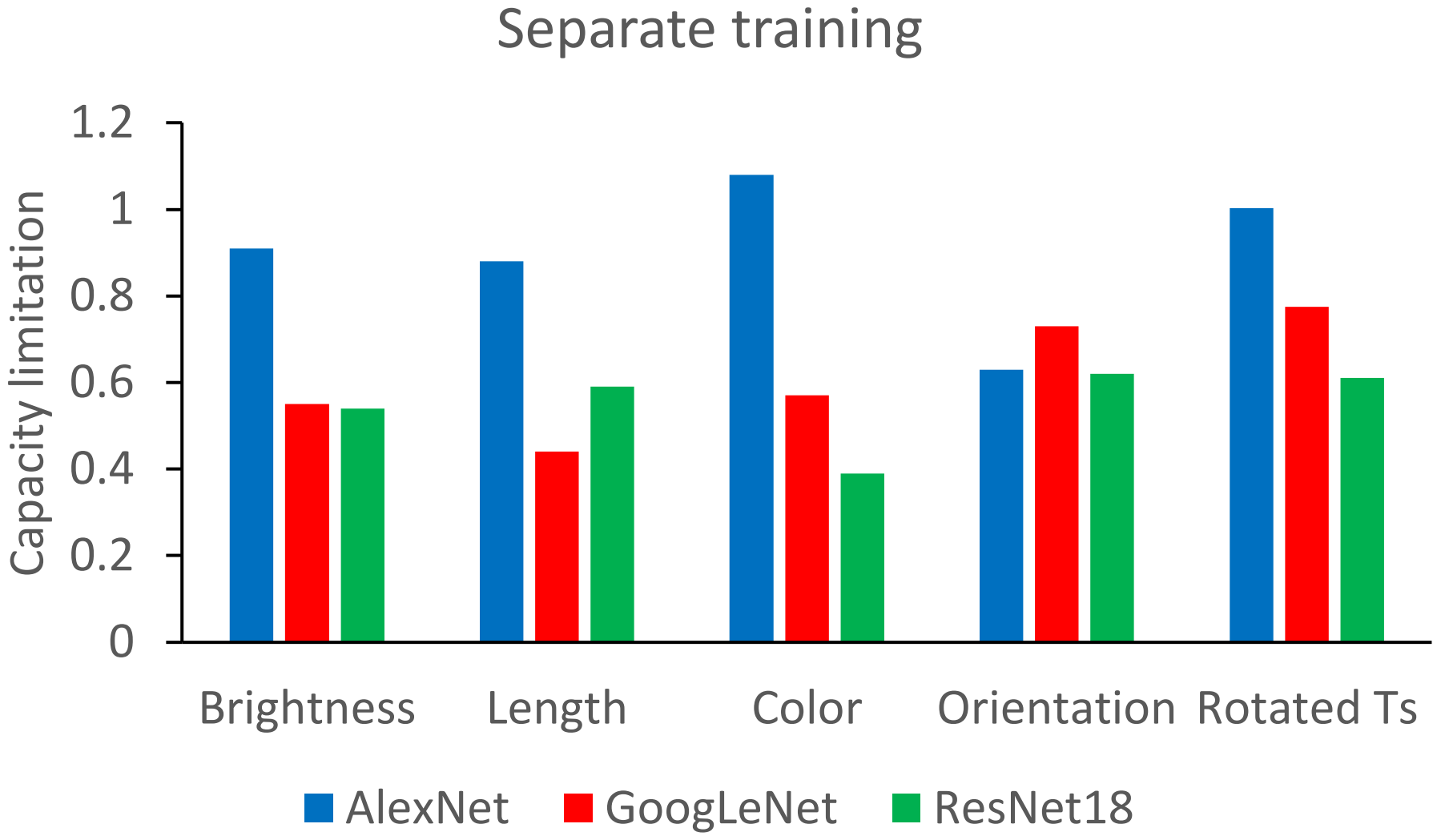

Figure 4. SDT based capacity limitation measures for different search experiments, different neural networks, and for two training procedures.

The somewhat smaller set size effects found with GoogLeNet and ResNet18 may indicate an effect of architectural innovations that supposedly increase sparsity of encoding and a better fit to biological vision (Szegedy et al., 2015). However, the present results are still far from unlimited capacity found in feature search with human observers.

Are there any theoretical ideas that could predict the set size effects from this study?
Some facts about linear networks may be useful. When the same set of units is used for a distributed coding of several variables, then interference from irrelevant ones appears as a noise in decoded pattern, with variance proportional to the number of irrelevant variables. This could explain an approximately square root decay of d prime as dependent on number of items. However, the square root decay is also consistent with a linear decision rule in multidimensional feature space, instead of the ideal decision rule.
I found some preference for SDT based search model over simple power function. However, the assumptions of this model may be incompatible with artificial neural networks. The idea of distinct representations of visual items, each with its own noise, looks too different from neural networks where representations are distributed over thousands of units, and noise is just an interference from other representations encoded by the same pool of units. The observed regularities reminiscent of ideal integration of noisy signals may have another explanation in these networks.

## Conclusions

This study revealed that deep convolutional neural networks, pretrained for object recognition, behave differently from human observers in classic visual search tasks. When human observers can search for a simple visual feature among homogeneous distractors very efficiently, artificial neural networks cannot. They exhibit comparable set size effects in both simple feature search and “complex” search for feature configurations. It is possible that training in object recognition does not build operations required for efficient search, and/or prewired architecture of traditional networks is not well suited for learning efficient visual search.

## References

Allport, D. A., Tipper, S. P., & Chmiel, N. R. J. (1985). Perceptual integration and post-categorical filtering. In M. I. Posner & O. S. M. Marin (Eds.), *Attention and performance XI* (pp. 107-132). Hillsdale, NJ: Erlbaum.

Bergen, J. R., & Julesz, B. (1983). Parallel versus serial processing in rapid pattern discrimination. *Nature, 303,* 696-698.

Broadbent, D. E. (1958). *Perception and communication*. London: Pergamon Press.

Cheal, M. L., Lyon, D. R., & Hubbard, D. C. (1991). Does attention have different effects on line orientation and line arrangement discrimination? *Quarterly Journal of Experimental Psychology, 43A*, 825-857.

Ciresan, D.C., Meier, U., Schmidhuber, J. (2012). Multi-column deep neural networks for image classification. In: IEEE Conference on Computer Vision and Pattern Recognition (CVPR 2012), 3642–3649.

Deutsch, J. A., & Deutsch, D. (1963). Attention: Some theoretical considerations. *Psychological Review, 70*, 80-90.

Estes, W. K., & Taylor, H. A. (1964). A detection method and probabilistic models for assessing information processing from brief visual displays. *Proceedings of the National Academy of Sciences of the United States of America, 52*, 446-454.

Geirhos, R., Temme, C. R. M., Rauber, J., Schütt, H. H., Bethge, M., & Wichmann, F. A. (2018). Generalization in humans and deep neural networks. *Advances in Neural Information Processing Systems 31*.

Khaligh-Razavi, S-M., & Kriegeskorte, N. (2014). Deep supervised, but not unsupervised, models may explain IT cortical representation. *PLoS Comput. Biol., 10*(11):e1003915.

Kinchla, R. A. (1974). Detecting target elements in multi-element arrays: A confusability model. *Perception and Psychophysics*, *15*, 149–158.

Krizhevsky A, Sutskever I, Hinton G (2012) ImageNet classification with deep convolutional neural networks. *Advances in Neural Information Processing Systems, 25*, 1097–1105.

Kubilius, J., Bracci, S., Op de Beeck, H. P. (2016) Deep neural networks as a computational model for Human shape sensitivity. *PLoS Comput. Biol., 12*(4): e1004896.

Lonnqvist, B., Clarke, A. D. F, & Chakravarthi, R. (2020). Crowding in humans is unlike that in convolutional neural networks. *Neural Networks, 126*, 262-274.

Mazyar, H., Van den Berg, R., & Ma, W. J. (2012). Does precision decrease with set size? *Journal of Vision, 12*(6).

Neisser, U. (1967). *Cognitive psychology*. New York: Appleton-Century-Crofts.

Nguyen A, Yosinski J, Clune J (2015) Deep neural networks are easily fooled: High confidence predictions for unrecognizable images. In: *Proceedings of the IEEE Conference on Computer Vision and Pattern Recognition*, pp 427–436.

Palmer, E. M., Fencsik, D.E., Flusberg, S.J., Horowitz, T.S., & Wolfe, J.M. (2011). Signal detection evidence for limited capacity in visual search. *Attention, Perception and Psychophysics, 73*, 2413-24.

Palmer, J. (1994). Set-size effects in visual search: The effect of attention is independent of the stimulus for simple tasks. *Vision Research, 34*, 1703-1721.

Palmer, J., Ames, C. T., & Lindsey, D. T. (1993). Measuring the effect of attention on simple visual search. *Journal of Experimental Psychology: Human, Perception and Performance, 19*, 108–130.

Palmer, J., Verghese, P., & Pavel, M. (2000). The psychophysics of visual search. *Vision Research 40*, 1227-1268.

Põder, E. (1999). Search for feature and for relative position: Measurement of capacity limitations. *Vision Research, 39*, 1321–1327.

Põder, E. (2017). Combining local and global limitations of visual search. *Journal of Vision, 17*(4): 10, 1-12.

Rosenholtz, R. (2017). Capacity limits and how the visual system copes with them. *Journal of Imaging Science and Technology* (Proc. HVEI, 2017).

Rosenholtz, R., Huang, J., & Ehinger, K.A. (2012). Rethinking the role of top-down attention in vision: effects attributable to a lossy representation in peripheral vision. *Frontiers in Psychology, 3:13,* 1-15.

Shaw, M. L. (1984). Division of attention among spatial locations: A fundamental difference between detection of letters and detection of luminance increments. In H. Bouma & D. G. Bouwhais (Eds.), *Attention and performance X* (pp. 109-121). Hillsdale, NJ: Erlbaum.

Simonyan, K. & Zisserman, A. (2014). Very deep convolutional networks for large-scale image recognition. arXiv Technical Report, arXiv:1409.1556.

Szegedy, C., Liu, W., Jia, Y., Sermanet, P., Reed, S., Anguelov, D., Erhan, D., Vanhoucke, V., & Rabinovich, A. (2015). Going deeper with convolutions. In *Proceedings of the IEEE Conference on Computer Vision and Pattern Recognition*, 1–9.

Treisman, A. M., Gelade, G. (1980). A feature integration theory of attention. *Cognitive Psychology, 12*, 97-136.

Wolfe, J. M. (1998). Visual search. In H. Pashler (Ed.), *Attention* (pp. 13–74). Hove, East Sussex, UK: Psychology Press Ltd.

Wolfe, J. M., & Bennett, S. C. (1996). Preattentive object files: Shapeless bundles of basic features. *Vision Research, 37*, 25-43.

Wolfe, J. M., Cave, K. R., & Franzel, S. L. (1989). Guided search: An alternative to the feature integration theory of attention. *Journal of Experimental Psychology: Human Perception and Performance 15*, 419-433.

Yamins, D. L. K., Hong, H., Cadieu, C. F., Solomon, E. A., Seibert, D., and DiCarlo, J. J. (2014). Performance-optimized hierarchical models predict neural responses in higher visual cortex. *PNAS, 111*, 8619-8624.